\pdfoutput=1

\documentclass[11pt]{article}

\usepackage[preprint]{acl}

\usepackage{times}
\usepackage{latexsym}

\usepackage[T1]{fontenc}

\usepackage[utf8]{inputenc}

\usepackage{microtype}

\usepackage{inconsolata}

\usepackage{graphicx}
\usepackage{caption}
\usepackage{subcaption}

\usepackage{xcolor}

\usepackage{amsthm}
\usepackage{algorithm}
\usepackage{algpseudocode}
\usepackage{adjustbox, booktabs}       
\usepackage{multirow}

\usepackage{amsmath,amsfonts,bm}

\def\eqref#1{equation~\ref{#1}}

\def\1{\bm{1}}

\DeclareMathAlphabet{\mathsfit}{\encodingdefault}{\sfdefault}{m}{sl}
\SetMathAlphabet{\mathsfit}{bold}{\encodingdefault}{\sfdefault}{bx}{n}

\newcommand{\R}{\mathbb{R}}

\newcommand\norm[1]{\Vert#1\Vert}

\newtheorem{theorem}{Theorem}

\usepackage[textsize=tiny]{todonotes}

\title{CompAct: Compressed Activations for Memory-Efficient LLM Training}

\author{
 \textbf{Yara Shamshoum\textsuperscript{*}},
 \textbf{Nitzan Hodos\textsuperscript{*}},
 \textbf{Yuval Sieradzki},
 \textbf{Assaf Schuster},
\\
 \textsuperscript{}Department of Computer Science, Technion - Israel Institute of Technology
\\
 \small{
   \href{mailto:email@domain}{\{yara-sh, hodosnitzan, syuvsier\}@campus.technion.ac.il}
 }
}

\begin{document}
\maketitle

\def\thefootnote{*}\footnotetext{Equal contribution.}

\begin{abstract}
We introduce CompAct, a technique that reduces peak memory utilization on GPU by 25-30\% for pretraining and 50\% for fine-tuning of LLMs.
Peak device memory is a major limiting factor in training LLMs, with various recent works aiming to reduce model memory. However most works don't target the largest component of allocated memory during training: the model's compute graph, which is stored for the backward pass.
By storing low-rank, compressed activations to be used in the backward pass we greatly reduce the required memory, unlike previous methods which only reduce optimizer overheads or the number of trained parameters.
Our compression uses random projection matrices, thus avoiding additional memory overheads.
Comparisons with previous techniques for either pretraining or fine-tuning show that CompAct substantially improves existing compute-performance tradeoffs.
We expect CompAct's savings to scale even higher for larger models.
\end{abstract}

\section{Introduction}
Training Large Language Models (LLMs) and fine-tuning them on downstream tasks has led to impressive results across various natural language applications \citep{t5, gpt3}. However, as LLMs scale from millions to hundreds of billions of parameters, the computational resources required for both pre-training and fine-tuning become prohibitive. 

While compute power is the primary bottleneck for those who train very large LLMs, memory requirements become the main limitation for researchers without access to vast hardware resources. This disparity severely limits the ability to advance the field of LLM training to only a select few.

\begin{figure}[!htbp]
  \centering
   \includegraphics[width=\linewidth]{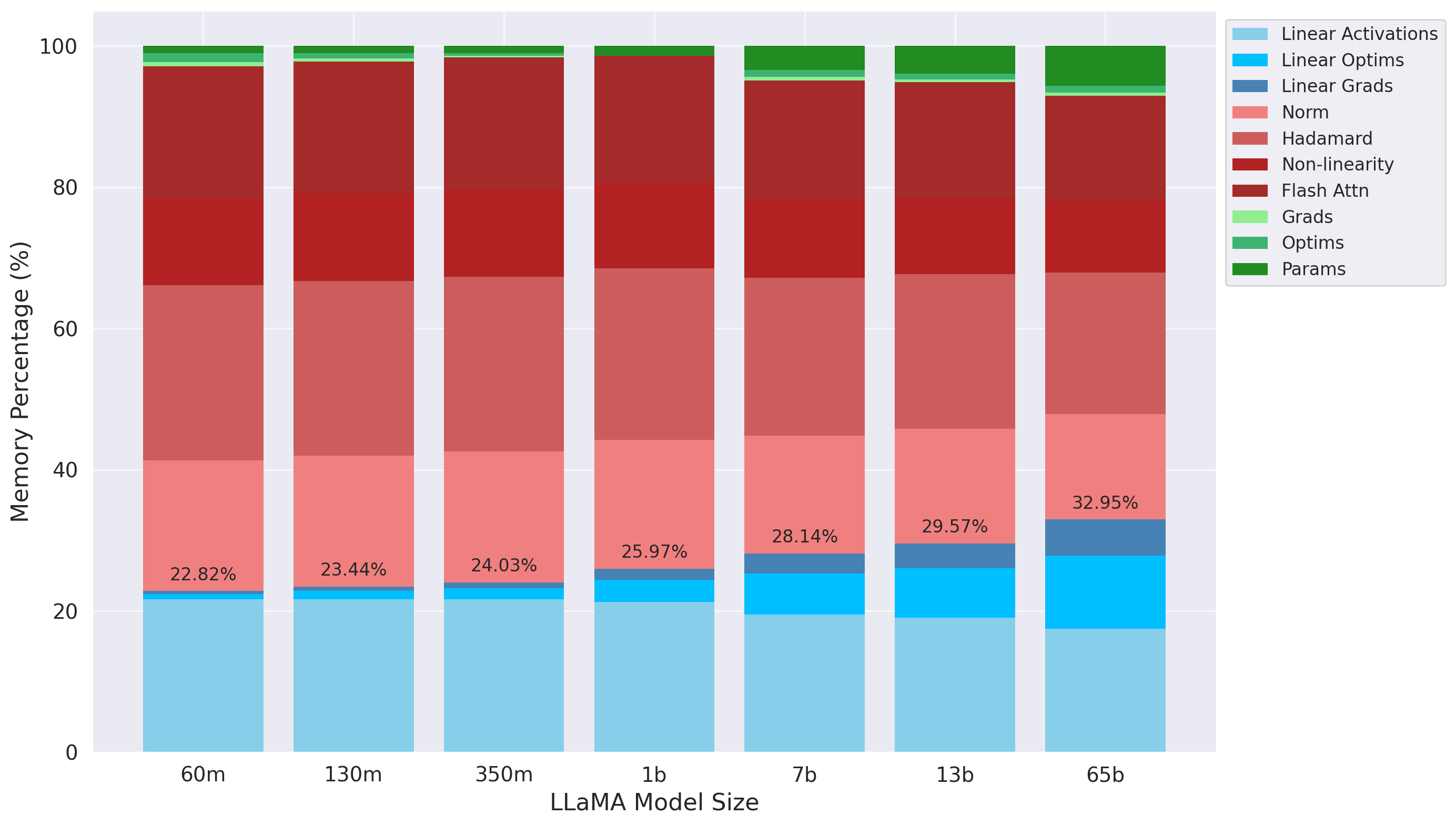}
    \caption{
    \textbf{Breakdown of memory components for various LLaMA model sizes,} with batch size $256$.
    \textcolor{blue}{\textbf{Blue}}: linear operations compressed by CompAct; \textcolor{red}{\textbf{Red}}: non-linear operations which CompAct doesn't compress; \textcolor{green}{\textbf{Green}}: model parameters and non-linear operation's optimizer states.
    Most of the memory is used by the computational graph. CompAct's compression gets more significant as model size increases, reaching almost 33\% for LLaMA 65B.
    With $r=n/8$, this translates to almost 30\% total memory saved.
    }
    \label{fig:mem-extension-demo}
\end{figure}

Recent works tackle memory reductions by applying a low rank approximation to model parameters \cite{lora, relora}, or to gradients after the backward pass \cite{grass, flora, galore}.
However, as seen in Figure \ref{fig:mem-extension-demo}, the main memory component is the computation graph itself. Its size also scales with batch size, in contrast with other memory components.
\begin{table*}
  \centering
  \begin{tabular}{lccccc}
    \toprule
     & \textbf{Original}& \textbf{LoRA} & \textbf{GaLore} & 
     \textbf{Flora} & \textbf{CompAct} \\
    \midrule
    \textbf{Weights} & $mn$  & $mr+nr$  & $mn$   & $mn$  & $mn$          \\
    \textbf{Gradients}            & $mn$ &  $mr+nr$   &$mn$ &$mn$  &$mr$           \\
    \textbf{Optim States} 
   & $2mn$ & $2(mr+nr)$  & $nr+2mr$    & $2mr$ & $2mr$ \\         
    \textbf{Activations}  
    &  $bln$   & $bln$  & $bln$     & $bln$  & $blr$\\
    \bottomrule
  \end{tabular}
  \caption{\textbf{Theoretical Memory Consumption by the different stages of the training pipeline}, assuming linear layers $W_t\in\mathbb{R}^{n\times m}$ and $m>n$. $b$ is batch size and $l$ is sequence length. $r$ is the dimensionality of the compressed activations and states.}
  \label{tab:theory_components}
\end{table*}

\begin{figure}[t]
  \includegraphics[clip, trim=4cm 2cm 4cm 2cm, width=\columnwidth]{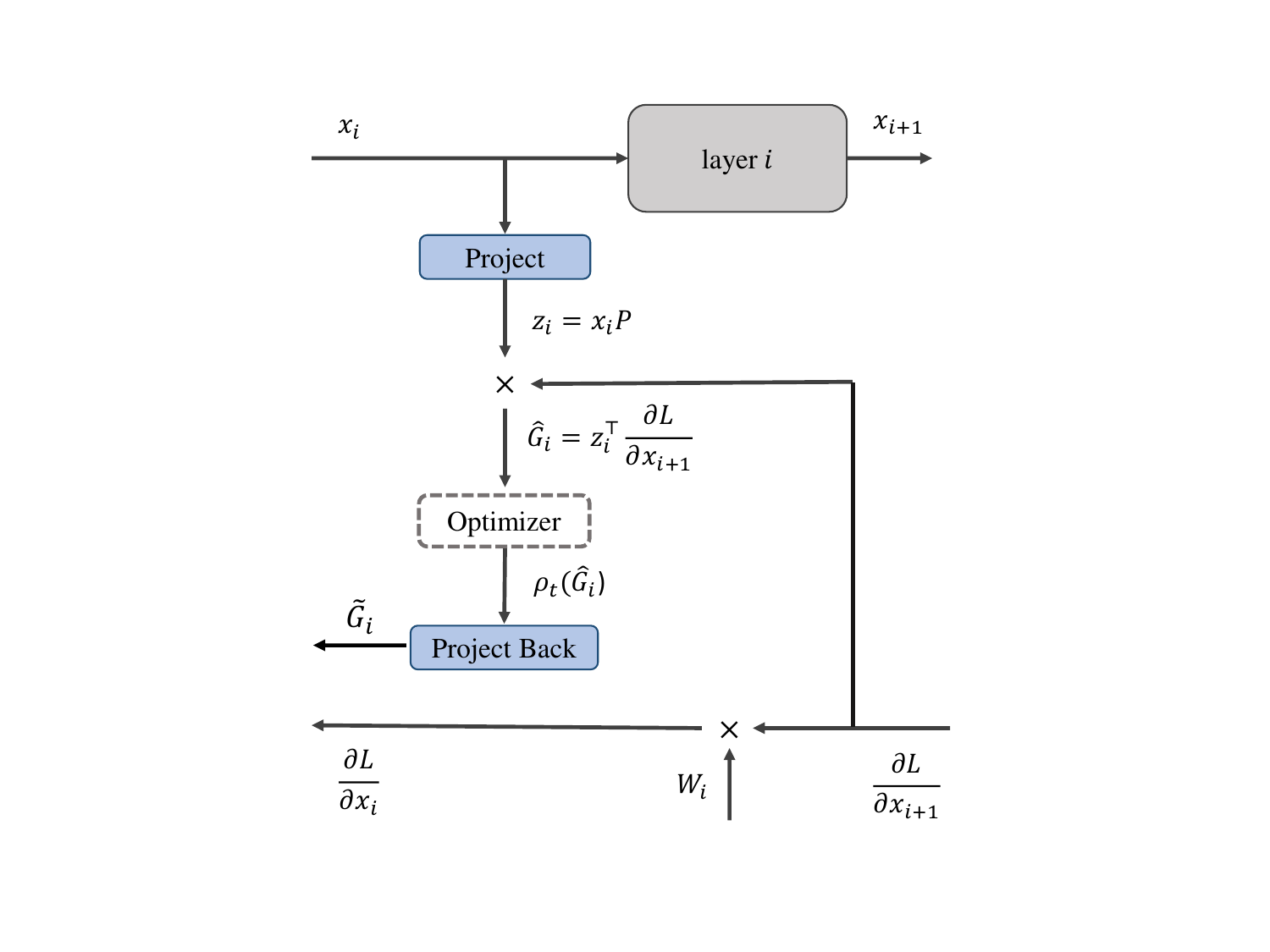}
  \caption{\textbf{Overview of CompAct.} 
  For a given linear layer $x_{i+1} = x_i W_{i+1}$, we project its input $x_i$ using a random projection matrix $P$, and save the result $z_{i}$ for the backward pass.
  During the backward pass, we first compute the compressed gradients $\hat{G_i}$ and update the optimizer's parameter update function $\rho_t(\hat{G_i})$. For Adam, $\rho_t$ represents gradient normalization using the first and second gradient moments. Finally, we decompress the gradient back to the full parameter size $\tilde{G_i}$ and perform an update step.
  }
  \label{fig:compact}
\end{figure}

In this work, we introduce CompAct, a novel optimization technique that saves low-rank-compressed activations during the forward pass, instead of the full activation tensors. Consequently, the resulting gradients are low-rank as well, also reducing the size of optimizer states. As CompAct decompresses the gradients back to full size only for the update step, it compresses a large part of the compute graph, which in turn translates to major memory savings.
Figure \ref{fig:compact} presents an overview of our method, while Table \ref{tab:theory_components} lists the scaling of different memory components, compared to other compression methods. 
CompAct is a logical next step from previous work, moving from low-rank parameters in \cite{lora}, through compressed low-rank gradients in \cite{galore}, to compressed activations.

Overall, CompAct achieves savings of about 17.3\% of \textit{peak device memory with practical batch sizes}, for pretraining LLaMA-350M \cite{llama}, and 50\% for fine-tuning RoBERTa-Base \cite{roberta}. As seen in Figure \ref{fig:mem-extension-demo}, the estimated memory reduction achieved by CompAct increases with model size. For LLaMA-65B, the estimated reduction is approximately 30\%. For LLaMA-350M, the estimated reduction is around 21\%, which is 4\% higher than the observed empirical value. This suggests a realistic memory saving range of 25\%-30\% for LLaMA-65B.

Choosing the low-rank projection used for compression is critical, as it can impact performance and reduce training throughput.
By using a random projection matrix sampled on the fly as in \cite{flora}, we eliminate the cost of computing and storing optimal projection matrices, which can be slow to compute and large enough to reduce our memory savings.


We show sound theoretical motivation justifying the use of random projections from recent works in Random Sketching theory \cite{meier2024fast}.
Finally, we present experimental measurements demonstrating that CompAct reduces memory significantly with minimal impact on model performance for both pretraining and finetuning.


\section{Related Work}
\paragraph{Memory-Efficient LLM Training} 
Memory-efficient methods have become crucial for training LLMs, especially during pretraining where the memory requirements for activations, gradients, and optimizer states are significant as these models scale. 

There are various approaches to making the training process more efficient, including mixed precision training \cite{mixed_precision}, quantization methods \cite{ACT,GACT,q_galore,qlora, coat}, parallelism techniques \cite{data_parallel_1, data_parallel_2, tensor_parallelism, pipeline_parallelism}, and low rank approximation methods. 
The latter being mostly unexplored for pretraining, it is the focus of this work.


\paragraph{Low Rank Approximation}

Prior works on efficient pretraining such as LoRA \cite{lora} have largely focused on low-rank parametrization techniques, where the model's weights \(W\) are decomposed into a product of two smaller matrices $W=BA$. 
While this method can reduce memory usage and computational cost, it often leads to performance degradation, as the low-rank constraint limits the network's ability to represent complex patterns. 

To address these limitations, approaches such as ReLoRA \cite{relora} and SLTrain \cite{sltrain} introduce more flexibility into the decomposition process, achieving a better balance between efficiency and performance. These methods go beyond strict low-rank parametrization by allowing for dynamic factorization, improving the network's expressiveness while retaining computational benefits.

\paragraph{Low Rank Gradient} 


A recent approach, GaLore \cite{galore} utilizes the low-rank property of gradients to learn a full-rank model efficiently, even during pretraining. 
Instead of approximating the model's weights, GaLore projects the gradients after the backward pass into a lower-dimensional subspace for the weight update, reducing the memory overhead without severely limiting the model's expressiveness.

As GaLore relies on periodic SVD computations to maintain adequate low-rank approximations of the gradients, which is very expensive in both time and memory, a variety of works focus on relieving this computational cost, either by strategically choosing the projection matrices \cite{subtrack}, quantizing them \cite{q_galore}, or replacing them by random projections \cite{grass,flora}.
However, these methods remain inapplicable when pretraining large models, as peak device memory is primarily determined by the activations stored on GPU (when the batch size is large) \cite{q_galore,grass}.

Essentially, compared to GaLore, our approach may be viewed as a change in the order of operations, applying the compression one step before GaLore does: when storing activations in memory for the backward pass, rather than to the gradients when updating the optimizer state.
As a result, CompAct satisfies their convergence theorem, which explains how it achieves comparable performance despite the drastic memory savings.

\paragraph{Activation Compression} 
Various works aim at reducing the memory cost of activations in deep learning in general. \cite{approx_bp} is a complementary work focusing on saving activation memory generated by nonlinear functions and normalization layers, whereas our work focuses on the activations generated by linear layers. 
The two methods can be combined to achieve even greater savings, although some adaptation is required.

VeLoRA \cite{velora} also aims to compress linear layer activations, however, they apply their paradigm to two specific layers of the model only, thus making their benefit marginal. 
We compare with their projection in our experimental section, see Section \ref{sec:ablation}. 
In any case, both \cite{velora} and \cite{approx_bp} remain unexplored for the setting of pretraining LLMs.

\paragraph{Activation Checkpointing}
CKPT \cite{CKPT}, also known as gradient checkpointing, reduces the memory footprint of the entire computation graph by saving the activations only at specific layers, or checkpoints.  
During backpropagation they recompute the forward pass between the current checkpoint and the previous one. 
The memory footprint of the entire compute graph can be reduced significantly, while incurring a 20\%-30\% compute cost overhead in most cases, as we empirically point out in Section \ref{sec:throughput}. 

\section{Method}
\subsection{Background}



Consider an input \(x\in \mathbb{R}^{b \cdot l\times n}\) where \(b\) is the batch size, \(l\) is the sequence length, and \(n\) is the number of input features. A linear layer with parameter \(W_t \in \R ^{n\times m}\) at learning step $t$ is applied as follows:
\begin{align}
    o = x W_t \in \R^{b \cdot l \times m},
\end{align}

where we eliminated the bias term for simplicity, as it is unaffected by the method.
During the forward pass, for each linear layer in the network, the input \(x\) is stored in memory at every intermediate layer. This is necessary for backpropagation, as it is used to compute the gradient of the weights using the chain rule:

\begin{align}
    G_t = \frac{\partial \mathcal{L}}{\partial W_t} = x^\top \frac{\partial \mathcal{L}}{\partial o} \in \R ^{n \times m}.
\end{align}

Once the gradient is computed, it is used to update the weights in the subsequent time step

\begin{align}
  W_{t+1} &= W_t -\eta \rho_t \left( \frac{\partial \mathcal{L}}{\partial W_t}\right).
\end{align}

Here, \(\eta\) represents the learning rate and \(\rho_t\) is an element-wise operation defined by the choice of optimizer, such as Adam.

Following the formulation in \citealp{galore}, GaLore projects the gradients into a lower-dimensional subspace before applying the optimizer, and projects them back to the original subspace for the weight update: 


\begin{align} \label{eq:galore}
    W_T &= W_0 + \eta \sum_{t=0}^{T-1} \Tilde{G}_t, \\
    \Tilde{G}_t &= P_t \rho_t(P_t^\top G_t Q_t)Q_t^\top .
\end{align}

Here \(P_t \in \R^{n\times r}\) and \(Q_t \in \R^{m\times r}\) are projection matrices, and \(\rho_t\) is the optimizer such as Adam.
In practice, to save memory, the projection is typically performed using only one of the two matrices, based on the smaller dimension between \(m\) and 
\(n\). This approach allows for efficient gradient compression and memory savings.
GaLore’s theoretical foundation, including its convergence properties, is captured in the following theorem:
\begin{theorem} \label{thm1}
    (Convergence of GaLore with fixed projections). Suppose the gradient follows the parametric form:
    \begin{align}
        G_t = \frac{1}{N} \sum_{i=1}^{N} (A_i -B_iW_tC_i)
    \end{align}
    with constant \(A_i\), PSD matrices \(B_i\) and \(C_i\) after \(t>t_0\), and \(A_i\), \(B_i\) and \(C_i\) have \(L_A\), \(L_B\) and \(L_C\) continuity with respect to \(W\) and \(\norm{W_t}\leq D\). Let \(R_t := P_t^\top G_t Q_t, \hat{B}_{it}:= P_t^\top B_i(W_t)P_t, \hat{C}_{it} := Q_t^\top C_i(W_t)Q_t \) and \(\kappa _t:= \frac{1}{N}\sum_i \lambda_{min}(\hat{B}_{it} \lambda_{min}\hat{C}_{it})\). If we choose constant \(P_t=P\) and \(Q_t=Q\), then GaLore with \(\rho_t=1\) satisfies:
    \begin{equation}
        \norm{R_t}_F \leq \left[1-\eta(\kappa_{r-1} - L_A - L_BL_CD^2\right]\norm{R_{t-1}}_
F
    \end{equation}
    As a result, if \(min_t \kappa_t > L_A + L_BL_CD^2, R_t\rightarrow 0\), and thus GaLore converges.
\end{theorem}

As stated in Theorem \ref{thm1}, the fastest convergence is achieved when projecting into a subspace corresponding to the largest eigenvalues of the matrices \(B_t, C_t\). To approximate this, GaLore employs a Singular Value Decomposition (SVD) on the gradient \(G_t\) every \(T\) timesteps to update the projection matrix. \(T\) is called the \emph{projection update period}.

Although this method reduces the memory cost of storing optimizer states, it introduces computational overhead due to the SVD calculation, and still requires saving the projection matrices in memory. The update period $T$ also creates a tradeoff between optimal model performance and training time, since for small $T$ the added SVD overhead becomes prohibitive, for large $T$ the projection might become stale and hurt model performance.




\subsection{CompAct}

An overview of the method is described in Algorithms \ref{alg:fw},\ref{alg:bw},\ref{alg:step}. 

To reduce memory usage during training, we propose saving a projected version of the input $z=xP\in\mathbb{R}^{b\cdot l\times r}$ during the forward pass, where $P\in\mathbb{R}^{n\times r}$ is a projection matrix that maps the input to a lower-dimensional subspace.
We choose $r$ to be a fraction of each layer's total dimensionality $n$, to achieve a consistent compression rate. Other works such as \cite{galore} chose the same $r$ for all compressed layers, which we think reduces potential compression gains.
In Section \ref{sec:experiments} we experiment with ratios $1/2,1/4,1/8$.

Using this low-rank projection $P$, the gradients with respect to the weights and the input are calculated as follows:
\begin{align}
  \hat{G}_t &= z^\top \frac{\partial \mathcal{L}}{\partial o} \in \R ^{r \times m}, \\
\frac{\partial \mathcal{L}}{\partial x} &= \frac{\partial \mathcal{L}}{\partial o} W_t ^\top \in \R ^{b \cdot l\times n}.
\end{align}
Our approach maintains the full forward pass, as well as the gradients with respect to the input.
However, the gradients with respect to the weights are computed within the reduced subspace. This means that the optimizer states are also maintained in this smaller subspace. Similar to \cite{galore},

\begin{align*}
    &M_t = \beta_1 M_{t-1} + (1-\beta_1) \hat G_t,\\
  &  V_t = \beta_2 V_{t-1} + (1-\beta_2)\hat G_t^2,\\
    &\rho_t(\hat G_t) = M_t / \sqrt{V_t+\epsilon}
\end{align*}

describes the Adam optimizer which we use in our analysis and experiments. Once the reduced gradient is obtained, we project it back to the original subspace for the full weight update using the same projection matrix $P$:

\begin{align}
  W_{t+1} &= W_t -\eta \tilde{G}_t,\\
  \tilde{G}_t &= \alpha P\rho_t (\hat{G}_t).
\end{align}
Where $\alpha$ is an optional gradient scaling constant.
By choosing $P$ such that $P P^\top \approx I$, $\hat{G}_t$ is a good approximation for the full gradient $G_t$.
This weight update is equivalent to GaLore’s (Equation \ref{eq:galore}) when \(Q=I\), so our method follows the convergence properties outlined in Theorem \ref{thm1}.


\begin{algorithm}
\caption{Forward Pass with CompAct}
\label{alg:fw}
\textbf{Input:} An input \(x\in \mathbb{R}^{b \cdot l\times n}\), a weight $W_t \in \R ^{n\times m}$, a layer seed $s\in\mathbb{N}$, a rank $r$.
\begin{algorithmic}[1]
    \State \textbf{set\_random\_seed}($s$)
    \State $P \leftarrow\mathcal{N}( 0, \frac{1}{r})\in \R^{n \times r}$
    \State  $o  \leftarrow x W_t \in \R^{b \cdot l \times m}$
    \State $z\leftarrow xP\in \R^{b \cdot l \times r}$
    \State  \textbf{save\_for\_backward}($z$, $W_t$) 
    \State \textbf{return} $o$
\end{algorithmic}
\end{algorithm}

\begin{algorithm}
\caption{Backward Pass with CompAct}
\label{alg:bw}
\textbf{Input:} An output gradient $\frac{\partial \mathcal{L}}{\partial o}\in \mathbb{R}^{b \cdot l\times m}$, A compressed activation $z\in\mathbb{R}^{b\cdot l\times r}$ a weight $W_t \in \R ^{n\times m}$.
\begin{algorithmic}[1]
    \State $\hat{G}_t \leftarrow z^\top \frac{\partial \mathcal{L}}{\partial o} \in \R ^{r \times m}$
    \State $\frac{\partial \mathcal{L}}{\partial x} \leftarrow \frac{\partial \mathcal{L}}{\partial o} W_t ^\top \in \R ^{b \cdot l\times n}$
    \State \textbf{return} $\frac{\partial \mathcal{L}}{\partial x}, \hat{G}_t $
\end{algorithmic}
\end{algorithm}

\subsection{Random Projection Matrix}
Choosing a data-dependent projection such as the SVD used by \cite{galore}, invariably forces extra compute to generate the projection, as well as memory allocation to store it for the backward pass.
Instead, following \cite{flora}, we opt for a random, data-independent projection which can be efficiently sampled on-the-fly and resampled for the backward pass by using the same seed as the forward pass.
We use a Gaussian Random Matrix, sampled independently for each layer with $\mu=0$ and $\sigma=1/r$: $P\in\R^{m\times r}, P_{ij}\sim\mathcal{N}(0, \frac{1}{r})$. Scaling the variance by $1/r$ ensures that $P P^\top\approx I$.

Using a random matrix from a Gaussian distribution ensures that the projected subspace maintains the norm of the original vectors with high probability \cite{jl_lemma_proof, gauss_preserves_norm}, which is critical for preserving information \cite{flora}.
Additionally, changing the projection every $T$ steps only required replacing the seed, which will not impact training throughput.

The following theorem by \cite{meier2024fast} demonstrates that training converges quickly with high probability, while using a Gaussian Random Matrix $P$:
 \begin{theorem}
     \textbf{(Sketching roughly preserves top singular values)}. Let \( P \in \mathbb{R}^{m \times r} \) have i.i.d entries sampled from \(\mathcal{N}(0, \frac{1}{r})\), and a low rank matrix \( A \in \mathbb{R}^{m \times n} \). We have:
     \begin{equation}
         \frac{\sigma_i(P^\top A)}{\sigma_i(A)} = \mathcal{O}(1).
     \end{equation}
     
\label{thm:singulars} \end{theorem}
Where $\sigma_i(M)$ is the $i$th largest singular value of matrix $M$.

Theorem \ref{thm:singulars} shows that the ratio of singular values \( \sigma_i(P^\top A)/\sigma_i(A) \) is fairly close to 1. As the main point in proving the convergence of GaLore is preserving the top singular values of the approximated matrix, this provides further motivation for why random sketching should work well.


\begin{algorithm}
\caption{Adam Update Step with CompAct}
\label{alg:step}
\textbf{Input:} A weight  $W_t \in \mathbb{R}^{n \times m}$, a compressed gradient $\hat{G}_t \in \mathbb{R}^{r \times m}$, a random seed $s\in\mathbb{N}$, Adam decay rates $\beta_1$, $\beta_2$, scale $\alpha$, learning rate $\eta$, rank $r$, projection update gap $T$. \\
Initialize Adam Moments $M_0, V_0 \in \mathbb{R}^{r \times m} \leftarrow 0,0$ \\
Initialize step $t \leftarrow 0$
\begin{algorithmic}[1]

\State $M_t \leftarrow \beta_1 \cdot M_{t-1} + (1 - \beta_1) \cdot \hat{G}_t$ 
\State $V_t \leftarrow \beta_2 \cdot V_{t-1} + (1 - \beta_2) \cdot \hat{G}_t^2$
\State $M_t \leftarrow M_t / (1 - \beta_1^t)$
\State $V_t \leftarrow V_t / (1 - \beta_2^t)$
\State $N_t \leftarrow M_t / (\sqrt{V_t} + \epsilon)$
\State \textbf{set\_random\_seed}($s$)
\State $P \leftarrow\mathcal{N}( 0, \frac{1}{r})$
\State $\tilde{G_t} \leftarrow 
\alpha \cdot P N_t$ \Comment{Project back to full dimension}
\State $W_{t+1} \leftarrow W_t - \eta \cdot \tilde{G_t}$

\State $t \leftarrow t + 1$
\If {$t \mod T = 0$}
    \State $s \leftarrow s + 1$ \Comment{Update Random Seed}
\EndIf
\State \textbf{return} $W_t$
\end{algorithmic}

\end{algorithm}

\section{Experiments}
\label{sec:experiments}
\begin{table*}
  \begin{adjustbox}{max width=\linewidth}
  \begin{tabular}{lcccccccc}
    \toprule
    Model size &  \multicolumn{2}{c}{\textbf{60M}} & \multicolumn{2}{c}{\textbf{130M}}  & \multicolumn{2}{c}{\textbf{350M}}  & \multicolumn{2}{c}{\textbf{1B}} \\
    \midrule
     & \textbf{Perplexity}  & \textbf{GPU} & \textbf{Perplexity} &  \textbf{GPU} &\textbf{Perplexity} & \textbf{GPU} &\textbf{Perplexity} & \textbf{GPU} \\
     &    & \textbf{Peak} &  &  \textbf{Peak} & & \textbf{Peak} & & \textbf{Peak} \\
    \midrule
    \textbf{Full-Rank}             & {34.06}  & {11.59} & {25.08} & {18.66} & {18.80}  & {39.97}   & {15.56} & {-}    \\
    \textbf{GaLore}                & {34.88}  & {11.56} &{25.36}  & {18.48} &{18.95} &    {39.24}   & {15.64} & {75.40}   \\
    \textbf{CompAct \(r=n/2\)}     & {32.78}  & {11.32} & {25.37}  & {17.97} & {19.26}   &  {37.94}   & {17.40} & {72.82}  \\
    \textbf{CompAct \(r=n/4\)}    & {34.41}  & {10.80} &{26.98}  & {16.78} &{20.45}  &   {34.71}  & {18.02} & {65.57}   \\
    \textbf{CompAct \(r=n/8\)}   & {36.42}  & {10.54} & {28.70}  & {16.19} & {21.91}  &   {33.03}    & {19.23} & {61.88}  \\
    \midrule
    Training Tokens                & \multicolumn{2}{c}{1.1B}   & \multicolumn{2}{c}{2.2B}    & \multicolumn{2}{c}{6.4B} & \multicolumn{2}{c}{13.1B}          \\
    \bottomrule
  \end{tabular}
  \end{adjustbox}
  \caption{
  \textbf{Pretraining perplexity and peak GPU memory for different model sizes and different training techniques.} Total training tokens are shown in the last row.
  As can be seen, CompAct reduces peak memory by up to 17\% for LLaMA 350M, with comparable perplexity to baseline.
  For larger model sizes we estimate the total memory saving to be roughly 30\%. The baseline for LLaMA 1B did not fit within the \(\sim 81\,\text{GB}\) memory available at the same batch size.
  }
  \label{tab:pretrain}
\end{table*}

In this section, we evaluate CompAct on both pretraining (Section \ref{sec:pretrain}) and finetuning tasks (Section \ref{sec:finetuning}). In all experiments, we apply CompAct to all attention and MLP blocks in all layers of the model, except for the output projection in the attention mechanism. For further details, see Appendix \ref{sec:appendix_pretrain}.
Moreover, we provide a comparison of CompAct's throughput and memory usage with other methods in Section \ref{sec:throughput}, and explore various types of projection matrices in Section \ref{sec:ablation}.

\subsection{Pretraining}
\label{sec:pretrain}

For pretraining, we apply CompAct to LLaMA-based models \cite{llama} of various sizes and train on the C4 (Colossal Clean Crawled Corpus) dataset, a commonly used dataset for training large-scale language models \cite{c4}. The models were trained without any data repetition.

Our experimental setup follows the methodology outlined in \cite{galore}, using a LLaMA-based architecture that includes RMSNorm and SwiGLU activations \cite{swiglu}. For each model size, we maintain the same set of hyperparameters, with the exception of the learning rate and the projection update gap which were tuned. Further details regarding the training setup and hyperparameters can be found in Appendix \ref{sec:appendix_pretrain}.

As shown in Table \ref{tab:pretrain}, CompAct achieves performance comparable to full-rank training, while displaying a superior performance-to-memory trade-off at smaller ranks, successfully decreasing the peak allocated GPU memory by 17\% in the largest model.

Additionally, We provide memory estimates of the various components for LLaMA 350M. As shown in Table \ref{tab:estimated_components}, CompAct's memory savings are substantial across all stages of the training process, with notable reductions in the memory required for activations, gradients, and optimizer states in the linear layers. These savings are critical, as they significantly lower the overall memory footprint during training, possibly enabling larger models or batch sizes to be processed within the same hardware constraints.

\begin{table}
  \begin{adjustbox}{center,max width=0.95\linewidth}
  \begin{tabular}{lccc}
    \toprule
     & \textbf{Original}& \textbf{GaLore} & 
     \textbf{CompAct} \\
    \midrule
    \textbf{Weights} 
    &  0.65GB& 0.65GB&0.65GB\\
    \textbf{Gradients} 
     &  0.65GB& 0.65GB&0.26GB\\
    \textbf{Optim States} 
    &  1.3GB& 0.54GB&0.52GB\\     
    \textbf{Activations}  
     &  7.0GB& 7.0GB&2.87GB\\
    \textbf{Peak Memory}  
    &  39.97GB& 39.21GB&34.71GB\\
    \bottomrule
  \end{tabular}
  \end{adjustbox}
  \caption{\textbf{Estimated GPU memory consumption by different components of the training pipeline of LLaMA 350M, along the measured peak allocated GPU Memory.} All methods share an additional constant of activations that are not linear, explaining the gap between the sum of parts and the peak memory. Galore utilizes $r=128$, while CompAct was measured with $r=n/4$. }
  \label{tab:estimated_components}
\end{table}

\subsection{Finetuning}
\label{sec:finetuning}
\begin{table*}
\begin{adjustbox}{max width=\linewidth}
  \begin{tabular}{c|c|cccccccc|c}
    \toprule
    \bf   & \bf Peak (MB) & \bf CoLA & \bf STS-B & \bf MRPC  & \bf  RTE  & \bf SST2 & \bf MNLI & \bf QNLI & \bf QQP & \bf Avg \\
    \midrule
      Full Fine-Tuning                      & 6298 & 62.24 & 90.92 & 91.30 & 79.42 & 94.57 & 87.18 & 92.33 & 92.28 & 86.28 \\
      \midrule
      GaLore ($r$=4)                       & 5816& 60.35 & 90.73 & 92.25 & 79.42 & 94.04 & 87.00 & 92.24 & 91.06 & 85.89 \\
      \textbf{CompAct (\(r\)=4)}   &  \textbf{3092} & 60.40& 90.61& 91.70& 76.17& 93.84&85.06& 91.70& 90.79& 85.03 \\
      \midrule
      GaLore ($r$=8)                       &  5819 & 60.06 & 90.82 & 92.01 & 79.78 & 94.38 & 87.17 & 92.20 & 91.11 & 85.94 \\
      \textbf{CompAct (\(r\)=8)}   & \textbf{3102} & 60.66& 90.57& 91.70& 76.90& 94.27&86.40& 92.70& 91.31 & 85.56 \\
    \bottomrule
  \end{tabular}
  \end{adjustbox}
  \caption{
  \textbf{Finetuning performance on several benchmarks for various compression rates with GaLore and CompAct.} We report the empirical mean of three runs of our approach per task. Peak Memory was measured on RTE task. 
  It is clear that both CompAct's and GaLore's performance is comparable with full finetuning, and very close to each other. However peak memory is vastly reduced with CompAct, with as much as 50\% total memory saved. See Appendix \ref{sec:appendix_finetune} for the more details.
  }
  \label{tab:finetune}
\end{table*}

We finetune the pretrained RoBERTa-base model \cite{roberta} on the GLUE benchmark, a widely used suite for evaluating NLP models across various tasks, including sentiment analysis and question answering  \cite{glue}. We apply CompAct and compared its performance to GaLore. Following the training setup and hyperparameters from GaLore, we only tuned the learning rate. More details can be found in Appendix \ref{sec:appendix_finetune}

As shown in Table \ref{tab:finetune}, CompAct achieves an extreme 50\% reduction in the peak allocated GPU memory while delivering comparable performance.

\subsection{Peak Memory and Throughput}
\label{sec:throughput}

Methods that primarily compress the optimizer states, such as GaLore, often need to be combined with other memory-saving techniques like activation checkpointing to achieve meaningful reductions in memory usage during training. However, activation checkpointing introduces additional computational overhead by requiring activations to be recomputed during the backward pass \cite{CKPT}, which can degrade training throughput. This trade-off means that while such methods may showcase memory benefits, they can negatively impact overall training efficiency.

  \begin{figure}[!htbp]
    \centering
    \begin{subfigure}{\linewidth}
        \centering
        \includegraphics[width=0.9\linewidth]{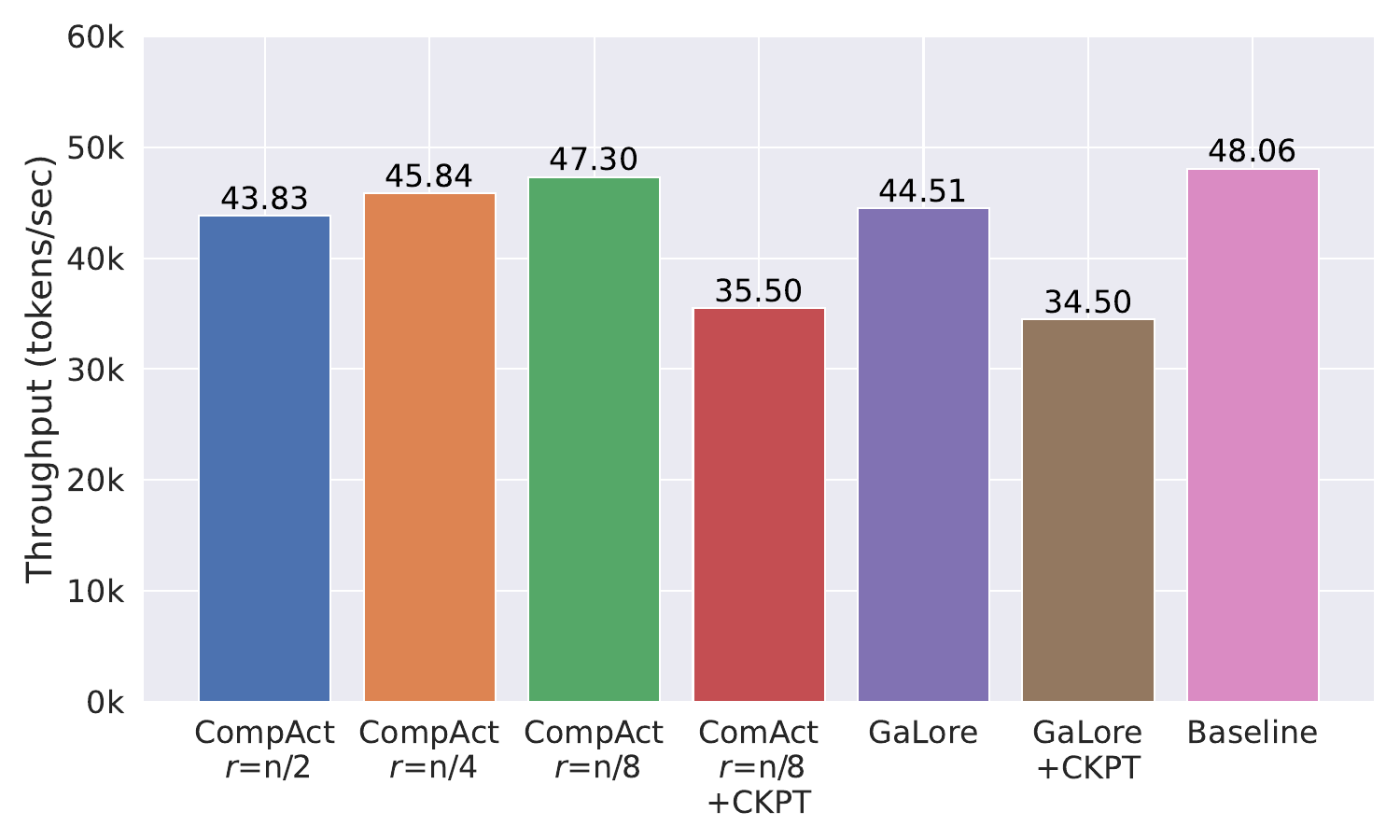}
        \label{fig:thpt} 
        \caption{}
    \end{subfigure}
    \hfill
    \begin{subfigure}{\linewidth}
        \centering
        \includegraphics[width=0.9\linewidth]{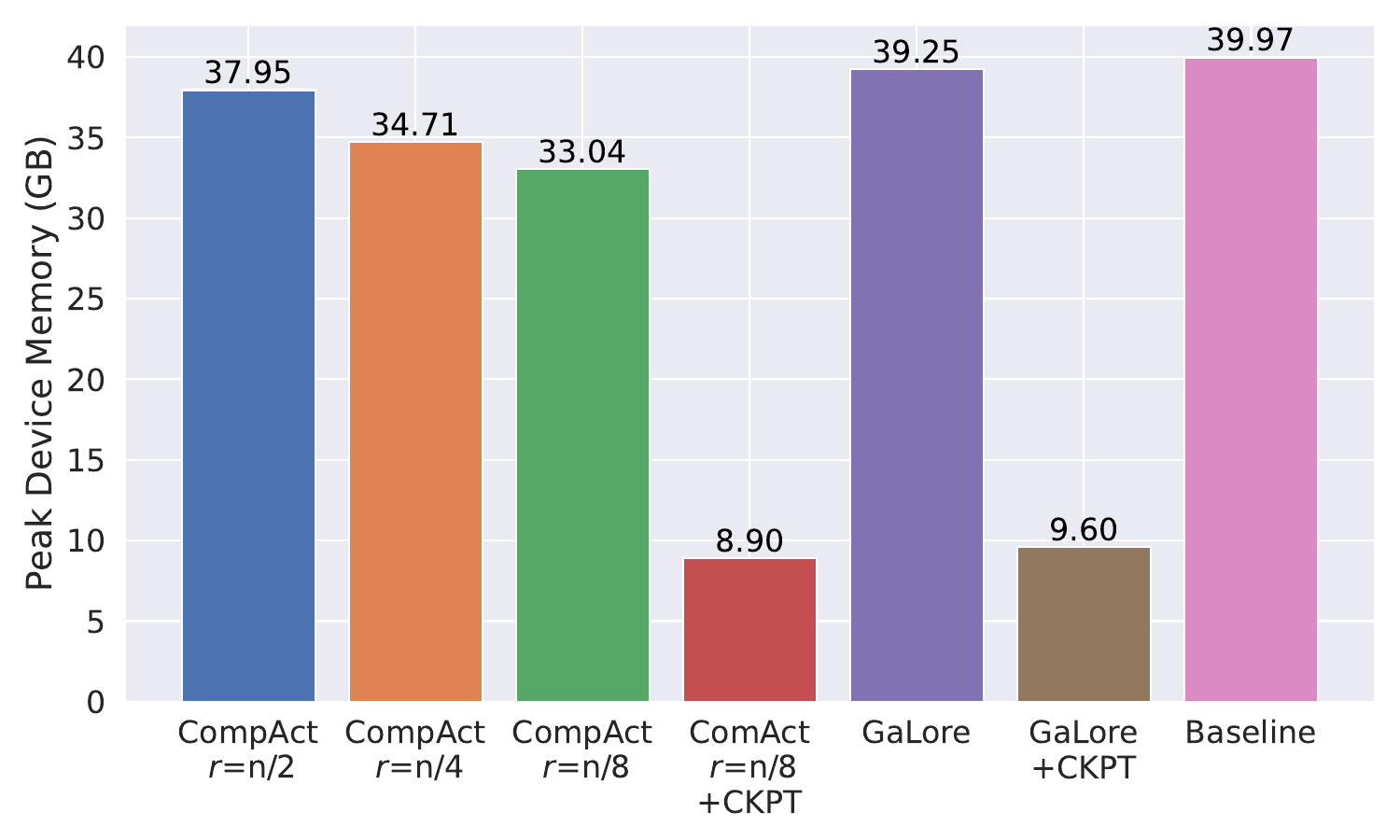}
        \label{fig:mem}
        \caption{}
    \end{subfigure}
    \caption{
    \textbf{(a) Throughput and (b) peak device memory during pretraining of LLaMa-350M.}
    As can be seen, using smaller ranks with CompAct achieves better compression than GaLore while increasing the throughput. When applying activation checkpointing (CKPT), CompAct remains competitive, achieving better throughput and a smaller memory footprint.
    }
    \label{fig:mem-thpt}
\end{figure}
  
We evaluate the throughput and memory peak of CompAct across various ranks and compare it against GaLore with and without activation checkpointing.
All experiments were conducted using LLaMA-350M with the same hyperparameters. For Galore, we utilized their official repository and adopted their optimal rank $r=256$ and projection update period $T=200$ for training this model.

Our results in Figure \ref{fig:mem-thpt} show that CompAct's reduction in peak GPU memory scales with $r$ as expected, reaching 17.3\% for $r=n/8$, while throughput also improves. This contrasts with the 1\% reduction achieved by standard GaLore, highlighting our assertion that optimizer state isn't a major contributor to total memory.

In both methods, applying activation checkpointing (CKPT) improves memory savings significantly while hurting total throughput. CompAct is still better than GaLore when using CKPT, though only slightly.

\section{Ablation Study}
\label{sec:ablation}
This section presents a series of ablation experiments examining how different design choices and hyperparameters affect training performance, memory usage, and convergence.

First, we explore the effects of different projection matrices on training performance when applied within the CompAct framework. We evaluate the following projection matrices:

\paragraph{Gaussian Projection} This is our primary method, where each layer samples a different Gaussian random matrix.

\paragraph{Gaussian Projection with Shared Seed} by setting the same random seed for all layers, we sample identical projection matrices for all layers (where dimensions permit). This investigates whether sharing the same subspace among different layers influences learning performance.

\paragraph{Sparse Johnson-Lindenstrauss (JL) Projection Matrices} JL matrices have guarantees for norm preservation, while being sparse. As shown in \cite{grass}, sparse operations can be highly efficient, and could point at future improvements for CompAct. We use the sparse JL matrix proposed in \cite{jl}.

\paragraph{VeLoRa Projection Matrices} VeLoRa \cite{velora} is, to our knowledge, the only other work that addresses the compression of activations of linear layers. However, their approach projects the activations back to the original space during the backward pass and computes the gradients in full rank. They also projected only the Down and Value layers of LLaMA, where CompAct applies to all linear layers. We employ their projection matrix within CompAct to evaluate its impact on our method.

We opted not to experiment with Singular Value Decomposition (SVD)-based projections in our method due to practical considerations. More on SVD in Compact is discussed in Appendix \ref{app:svd}

For each type of projection matrix, we train LLaMA-60M with rank \(r=n/4\). We conducted experiments using learning rates from [\(1e-2, 5e-3, 1e-3\)]. All other hyperparameters were identical to those used in Section \ref{sec:pretrain}.



\begin{figure}[!htbp]
    \centering
    \begin{subfigure}{\linewidth}
        \centering
        \includegraphics[clip, trim=0.5cm 0cm 0.5cm 0.5cm, width=\columnwidth]{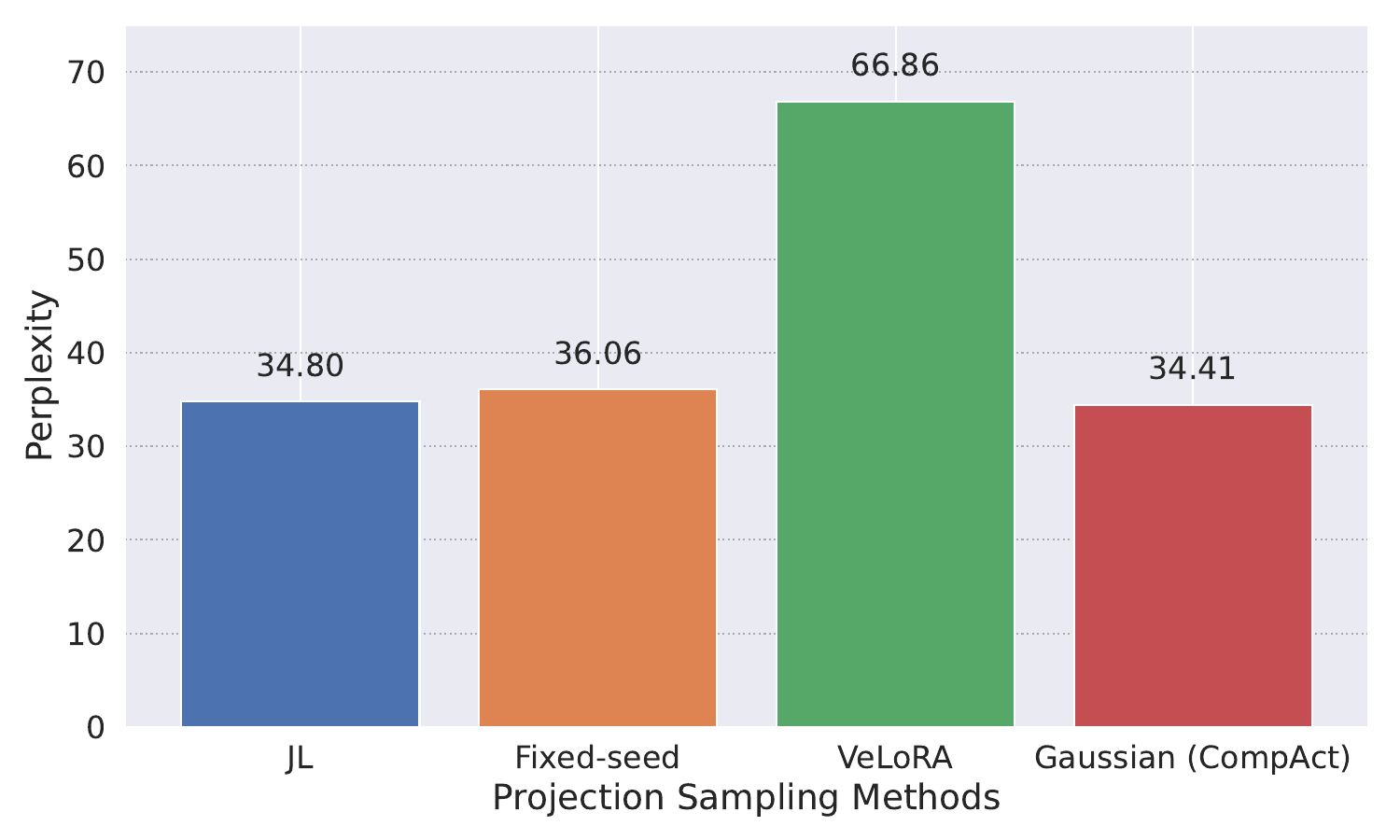}
        \label{fig:ab-perp}
    \end{subfigure}
    \caption{
     \textbf{Final model perplexity of CompAct with $r=n/4$ for different choices of projection matrices.}
     Both Gaussian seed choices and the JL projection achieve comparable results.
    }
    \label{fig:ablations}
\end{figure}

As shown in  Figure \ref{fig:ablations}, using Gaussian projections with different seeds per layer slightly improved performance compared to a shared seed, suggesting that utilizing different subspaces for different layers enhances the learning capacity of the model. Additionally, the sparse JL projections performed comparably to the dense Gaussian projections. This is a promising result, suggesting the viability of efficient sparse operations to further improve the benefits of CompAct. Finally, incorporating the projection matrix from VeLoRa into CompAct performed poorly. This can be attributed to differences in how VeLoRa handles the backward pass, by projecting activations back and computing full-rank gradients. This gap is somewhat expected, as they only used their projection on two types of layers, whereas we applied the compression more broadly. For the loss curves, see Appendix \ref{sec:appendix_c}.

\begin{figure}[!htbp]
    \centering
        \includegraphics[width=\columnwidth]{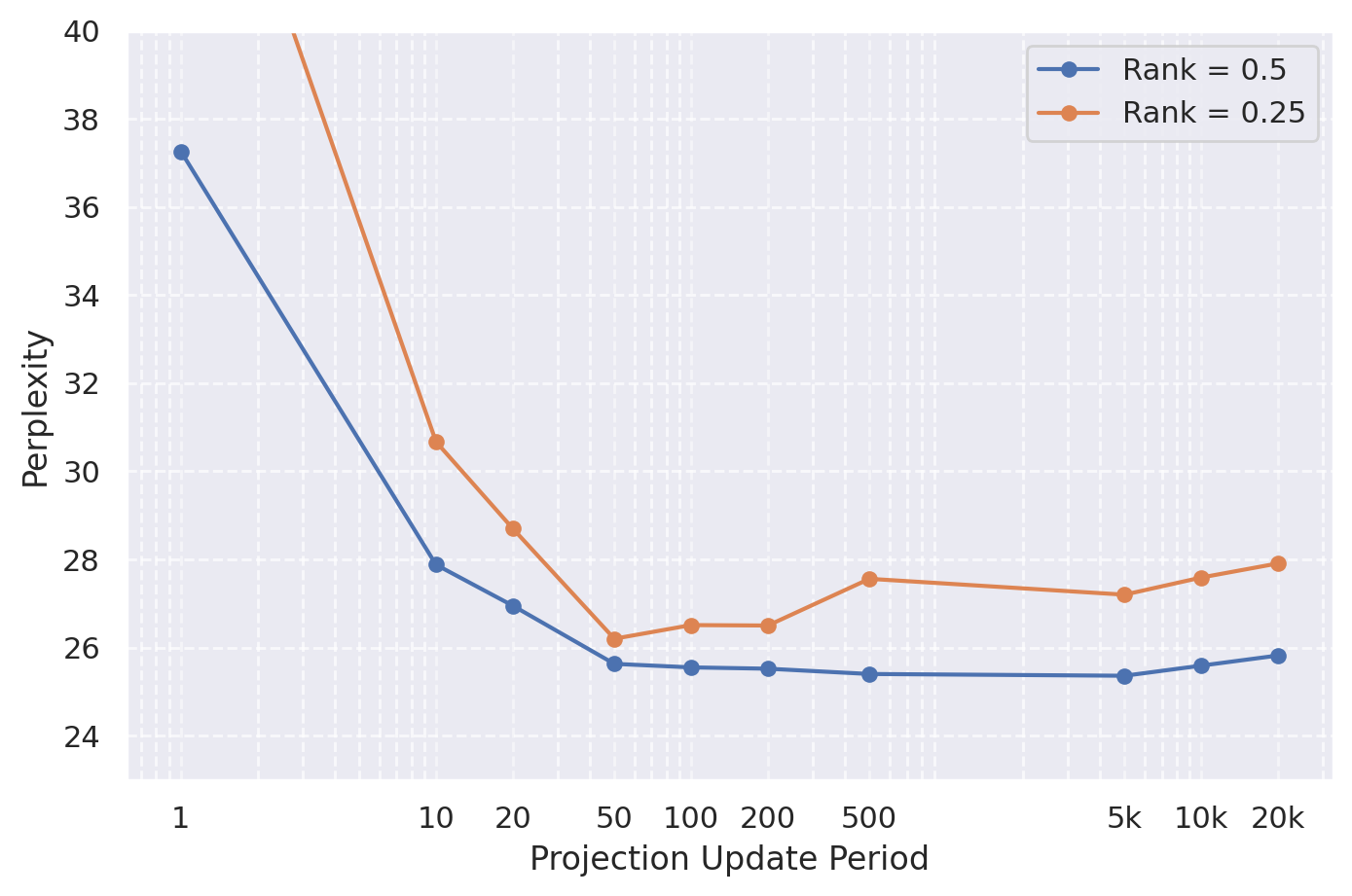}
    \caption{
     \textbf{Ablation on LlaMA 130M} -  Effect of varying projection update periods \(T\) on performance across different ranks in CompAct.}
    \label{fig:gap-ablations}
\end{figure}

Next, we examine how different training hyperparameters impact model convergence and memory efficiency.

\paragraph{Projection Update Period \(T\):}  We first analyze the influence of the projection update period \(T\) on the convergence of of LLaMA-130M when trained with CompAct.To do so, we conduct experiments with varying update intervals. A learning rate of \(5e-3\) is used by default, but if training becomes unstable at a given \(T\), we reduce the learning rate until stability is achieved.\\
As shown in Figure \ref{fig:gap-ablations}, an optimal \(T\) range emerges: updating the projection matrix too frequently or too infrequently both slow down convergence. This trend remains consistent across the couple of ranks we show.

\begin{figure}[!htbp]
    \centering
        \includegraphics[width=\columnwidth]{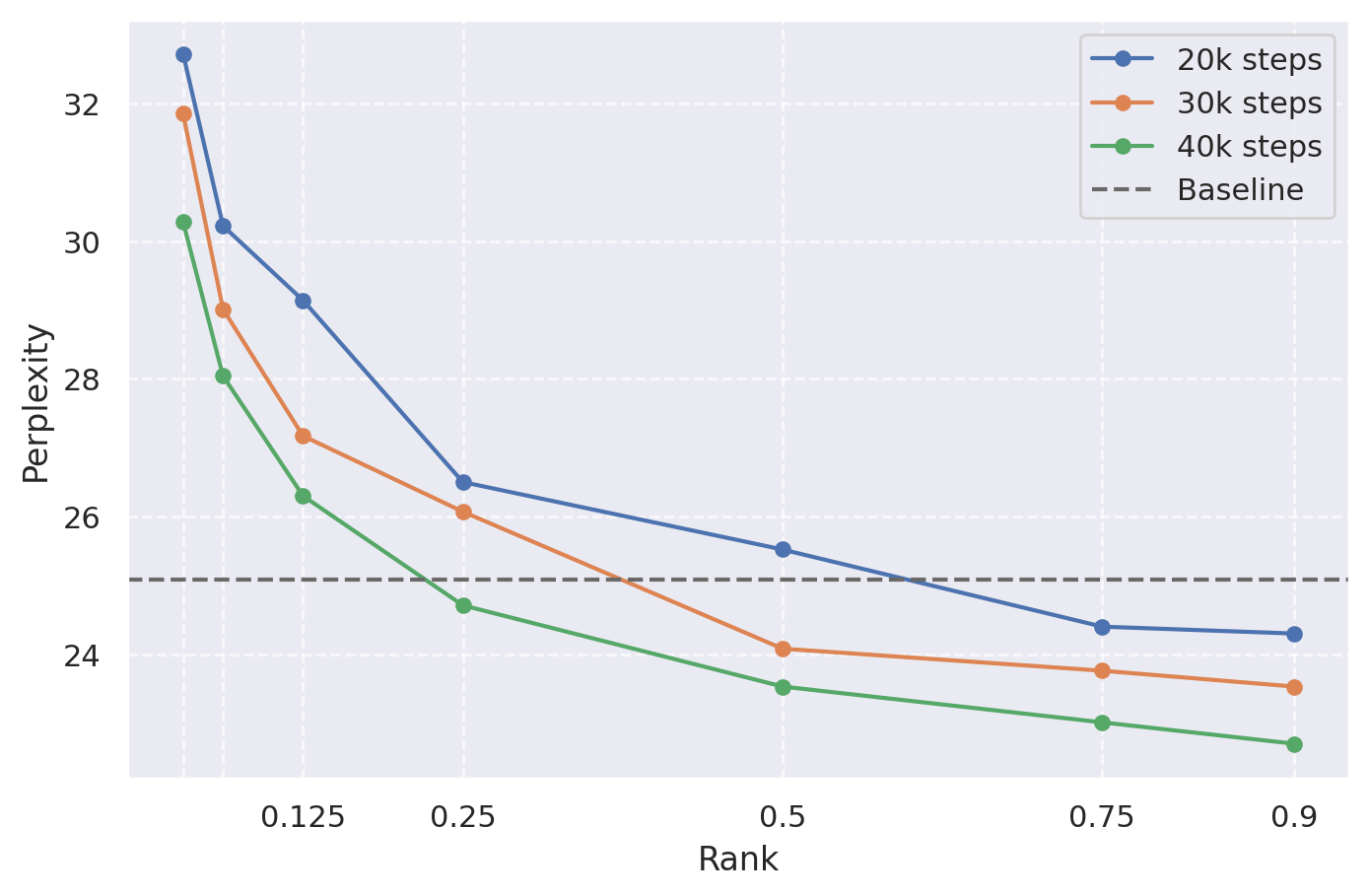}
    \caption{
     \textbf{Perplexity vs. Rank} - Effect of different ranks on the performance of CompAct-trained LLaMa-130M across varying training steps.
      The baseline (no compression) is trained for 20K steps.}
    \label{fig:rank-ablations}
\end{figure}

\paragraph{Rank and Training Steps:} Next, we investigate the effect of rank and training duration on model convergence.Here, LLaMA-130M is trained with a projection update period of \(T=200\) and an initial learning rate of \(5e-3\), which is reduced to \(4e-3\) if instability occurs.\\
Figure \ref{fig:rank-ablations} shows the impact of varying the rank of the projection matrix over different training durations. As expected, lower ranks lead to more performance degradation, while increasing the rank improves model performance. Additionally, for any given rank, training for more steps yields better results.\\
Crucially, larger ranks require fewer training steps to match or surpass the baseline, whereas lower ranks need extended training to compensate for their reduced expressivity. This highlights a trade-off between memory, training time, and final performance. For instance, when constrained by a small GPU, using a more aggressive compression (e.g., rank 0.25) allows training to fit within memory limits. Depending on the number of training steps available, this trade-off can still yield competitive performance, and with sufficient steps, even exceed the baseline.

\begin{figure}[!htbp]
    \centering
        \includegraphics[width=\columnwidth]{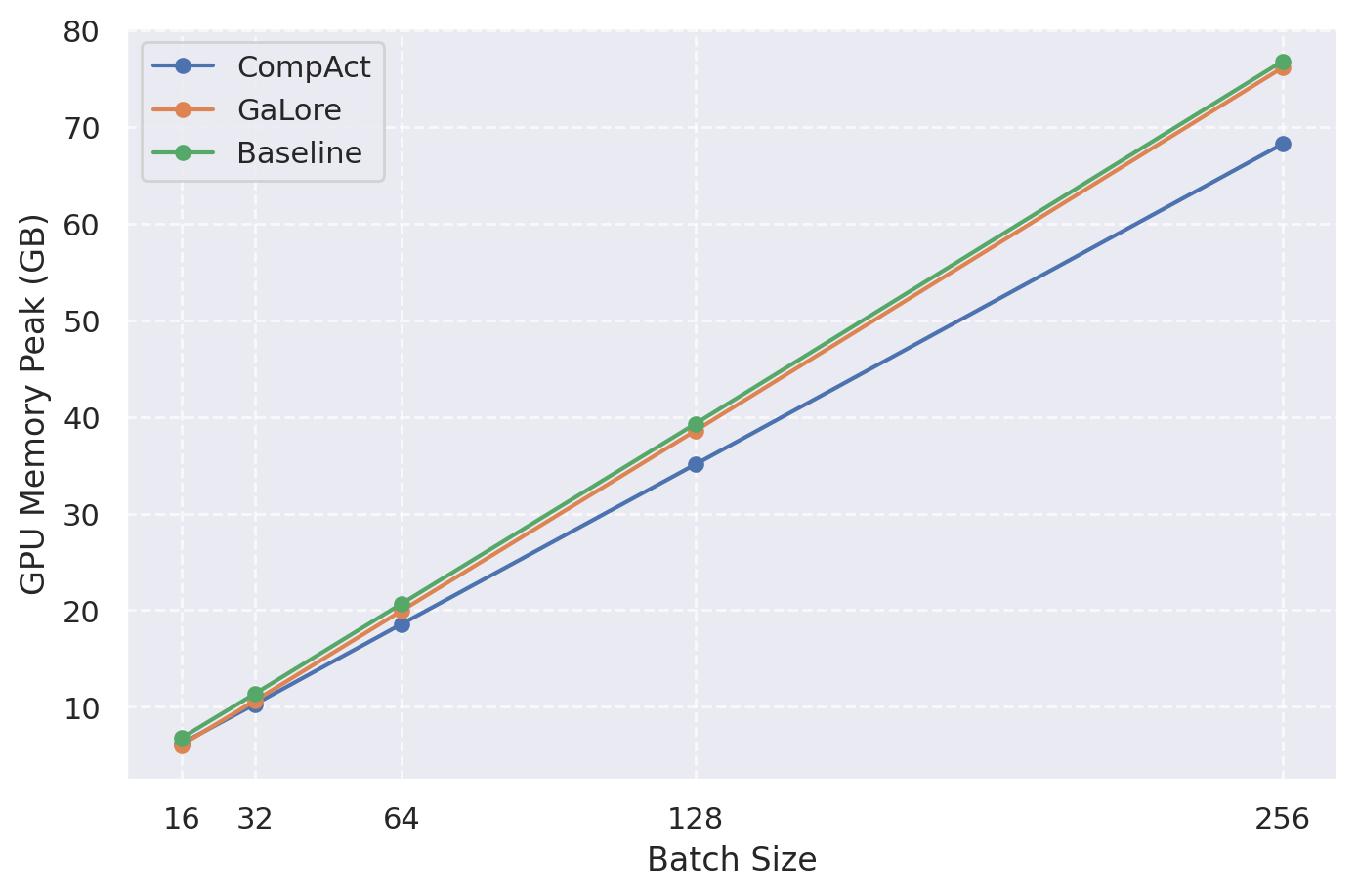}
    \caption{
  \textbf{Peak GPU memory vs. Batch Size.}
  Peak GPU memory usage of LLaMa-350M for different batch sizes is shown for CompAct, GaLore, and the baseline.
}\label{fig:batch-ablation}
\end{figure}

\paragraph{Training Batch Size:} Finally, we examine how batch size affects peak GPU memory usage, demonstrating the benefits of CompAct when handling varying activation sizes. We measure the peak GPU memory consumption while training LLaMA-350M without gradient accumulation across different batch sizes. The comparison includes the baseline model, GaLore (with a rank of 256), and CompAct (with a rank of 0.25).\\
As shown in Figure \ref{fig:batch-ablation}, CompAct significantly reduces peak memory usage, and its benefit scales with batch size, whereas GaLore provides a fixed reduction in peak memory relative to the baseline. This difference arises because GaLore primarily compresses optimizer states, which are independent of batch size, while CompAct reduces the memory footprint of activations, which grow with batch size.\\
This result is particularly important because larger batch sizes are known to improve training throughput\cite{tempo}. By lowering peak memory requirements, CompAct enables the use of larger batches, potentially increasing training efficiency without exceeding hardware constraints.

\section{Conclusion}
In this work, we presented CompAct, a memory-efficient method for training LLMs by compressing activations, gradients, and optimizer states of linear layers. We demonstrate that CompAct achieves significant memory savings for training LLMs, reaching 25\%-30\% memory reduction for pretraining LLaMA-65B and 50\% for RoBERTa-Base, with minimal impact on training throughput and performance.
Our method is easily scalable, applicable to various model sizes, and is easily composable with other techniques.

By directly compressing the compute graph during training, CompAct targets a major component of peak device memory which was neglected in recent works. We believe this approach should guide future work for further memory gains.
A good example could be incorporating sparse random projections into CompAct, which would reduce the computational cost associated with sampling and matrix operations. Another area for improvement is the approximation of intermediate activations, such as those generated by FlashAttention, Hadamard products, and non-linearities, which require significant memory. By addressing these memory-intensive operations, CompAct’s memory reductions can be extended even further.

\section{Limitations}
While CompAct offers significant memory savings and maintains high throughput, a few limitations should be noted.

First, although using Gaussian random matrices allows for on-the-fly sampling and eliminates the memory overhead of storing projection matrices, it can introduce some computational overhead due to frequent sampling and multiplications. A possible solution is to replace them with sparse random projections. These could not only reduce the computational cost but also improve throughput by fusing the sampling and multiplication steps, potentially outperforming the current baseline.

Another limitation of CompAct is that it currently focuses on compressing linear layers, leaving other memory-intensive operations, such as FlashAttention and Hadamard products, uncompressed. These operations consume substantial memory, and future work could explore compressing their activations directly within the computation graph without compromising model performance.

Finally, while we demonstrated that CompAct provides additional memory savings when combined with activation checkpointing—by avoiding the need to store full gradients in memory—the integration could be further optimized. Recomputing the activation compression during the backward pass could reduce the overhead introduced by checkpointing and improve throughput. Integrating CompAct with methods such as those proposed in \cite{approx_bp} could further smooth this process and enhance training efficiency.

\bibliography{acl_latex}

\begin{thebibliography}{32}
\providecommand{\natexlab}[1]{#1}

\bibitem[{Anonymous(2024{\natexlab{a}})}]{coat}
Anonymous. 2024{\natexlab{a}}.
\newblock \href {https://openreview.net/forum?id=XfKSDgqIRj} {{COAT}: Compressing optimizer states and activations for memory-efficient {FP}8 training}.
\newblock In \emph{Submitted to The Thirteenth International Conference on Learning Representations}.
\newblock Under review.

\bibitem[{Anonymous(2024{\natexlab{b}})}]{q_galore}
Anonymous. 2024{\natexlab{b}}.
\newblock \href {https://openreview.net/forum?id=rBzvEEbrF7} {Q-galore: Quantized galore with {INT}4 projection and layer-adaptive low-rank gradients}.
\newblock In \emph{Submitted to The Thirteenth International Conference on Learning Representations}.
\newblock Under review.

\bibitem[{Anonymous(2024{\natexlab{c}})}]{subtrack}
Anonymous. 2024{\natexlab{c}}.
\newblock \href {https://openreview.net/forum?id=nR0n4R1Ck2} {Subtrack your grad: Gradient subspace tracking for memory-efficient {LLM} training and fine-tuning}.
\newblock In \emph{Submitted to The Thirteenth International Conference on Learning Representations}.
\newblock Under review.

\bibitem[{Brown et~al.(2020)Brown, Mann, Ryder, Subbiah, Kaplan, Dhariwal, Neelakantan, Shyam, Sastry, Askell, Agarwal, Herbert-Voss, Krueger, Henighan, Child, Ramesh, Ziegler, Wu, Winter, Hesse, Chen, Sigler, Litwin, Gray, Chess, Clark, Berner, McCandlish, Radford, Sutskever, and Amodei}]{gpt3}
Tom~B. Brown, Benjamin Mann, Nick Ryder, Melanie Subbiah, Jared Kaplan, Prafulla Dhariwal, Arvind Neelakantan, Pranav Shyam, Girish Sastry, Amanda Askell, Sandhini Agarwal, Ariel Herbert-Voss, Gretchen Krueger, Tom Henighan, Rewon Child, Aditya Ramesh, Daniel~M. Ziegler, Jeffrey Wu, Clemens Winter, Christopher Hesse, Mark Chen, Eric Sigler, Mateusz Litwin, Scott Gray, Benjamin Chess, Jack Clark, Christopher Berner, Sam McCandlish, Alec Radford, Ilya Sutskever, and Dario Amodei. 2020.
\newblock \href {https://arxiv.org/abs/2005.14165} {Language models are few-shot learners}.
\newblock \emph{Preprint}, arXiv:2005.14165.

\bibitem[{Chen et~al.(2016)Chen, Xu, Zhang, and Guestrin}]{CKPT}
Tianqi Chen, Bing Xu, Chiyuan Zhang, and Carlos Guestrin. 2016.
\newblock \href {https://arxiv.org/abs/1604.06174} {Training deep nets with sublinear memory cost}.
\newblock \emph{Preprint}, arXiv:1604.06174.

\bibitem[{Dao et~al.(2022)Dao, Fu, Ermon, Rudra, and Ré}]{flash_attention}
Tri Dao, Daniel~Y. Fu, Stefano Ermon, Atri Rudra, and Christopher Ré. 2022.
\newblock \href {https://arxiv.org/abs/2205.14135} {Flashattention: Fast and memory-efficient exact attention with io-awareness}.
\newblock \emph{Preprint}, arXiv:2205.14135.

\bibitem[{Dasgupta et~al.(2010)Dasgupta, Kumar, and Sarl{\'{o}}s}]{jl}
Anirban Dasgupta, Ravi Kumar, and Tam{\'{a}}s Sarl{\'{o}}s. 2010.
\newblock \href {https://arxiv.org/abs/1004.4240} {A sparse johnson--lindenstrauss transform}.
\newblock \emph{CoRR}, abs/1004.4240.

\bibitem[{Dasgupta and Gupta(2003)}]{jl_lemma_proof}
Sanjoy Dasgupta and Anupam Gupta. 2003.
\newblock \href {https://doi.org/10.1002/rsa.10073} {An elementary proof of a theorem of johnson and lindenstrauss}.
\newblock \emph{Random Structures \& Algorithms}, 22(1):60--65.

\bibitem[{Dean et~al.(2012)Dean, Corrado, Monga, Chen, Devin, Mao, Ranzato, Senior, Tucker, Yang, Le, and Ng}]{data_parallel_1}
Jeffrey Dean, Greg Corrado, Rajat Monga, Kai Chen, Matthieu Devin, Mark Mao, Marc\textquotesingle~aurelio Ranzato, Andrew Senior, Paul Tucker, Ke~Yang, Quoc Le, and Andrew Ng. 2012.
\newblock \href {https://proceedings.neurips.cc/paper_files/paper/2012/file/6aca97005c68f1206823815f66102863-Paper.pdf} {Large scale distributed deep networks}.
\newblock In \emph{Advances in Neural Information Processing Systems}, volume~25. Curran Associates, Inc.

\bibitem[{Dettmers et~al.(2023)Dettmers, Pagnoni, Holtzman, and Zettlemoyer}]{qlora}
Tim Dettmers, Artidoro Pagnoni, Ari Holtzman, and Luke Zettlemoyer. 2023.
\newblock \href {https://arxiv.org/abs/2305.14314} {Qlora: Efficient finetuning of quantized llms}.
\newblock \emph{Preprint}, arXiv:2305.14314.

\bibitem[{Han et~al.(2024)Han, Li, Huang, Hong, Takeda, Jawanpuria, and Mishra}]{sltrain}
Andi Han, Jiaxiang Li, Wei Huang, Mingyi Hong, Akiko Takeda, Pratik Jawanpuria, and Bamdev Mishra. 2024.
\newblock \href {https://arxiv.org/abs/2406.02214} {Sltrain: a sparse plus low-rank approach for parameter and memory efficient pretraining}.
\newblock \emph{Preprint}, arXiv:2406.02214.

\bibitem[{Hao et~al.(2024)Hao, Cao, and Mou}]{flora}
Yongchang Hao, Yanshuai Cao, and Lili Mou. 2024.
\newblock \href {https://arxiv.org/abs/2402.03293} {Flora: Low-rank adapters are secretly gradient compressors}.
\newblock \emph{Preprint}, arXiv:2402.03293.

\bibitem[{Hu et~al.(2021)Hu, Shen, Wallis, Allen-Zhu, Li, Wang, Wang, and Chen}]{lora}
Edward~J. Hu, Yelong Shen, Phillip Wallis, Zeyuan Allen-Zhu, Yuanzhi Li, Shean Wang, Lu~Wang, and Weizhu Chen. 2021.
\newblock \href {https://arxiv.org/abs/2106.09685} {Lora: Low-rank adaptation of large language models}.
\newblock \emph{Preprint}, arXiv:2106.09685.

\bibitem[{Huang et~al.(2019)Huang, Cheng, Bapna, Firat, Chen, Chen, Lee, Ngiam, Le, Wu, and Chen}]{pipeline_parallelism}
Yanping Huang, Youlong Cheng, Ankur Bapna, Orhan Firat, Dehao Chen, Mia Chen, HyoukJoong Lee, Jiquan Ngiam, Quoc~V Le, Yonghui Wu, and zhifeng Chen. 2019.
\newblock \href {https://proceedings.neurips.cc/paper_files/paper/2019/file/093f65e080a295f8076b1c5722a46aa2-Paper.pdf} {Gpipe: Efficient training of giant neural networks using pipeline parallelism}.
\newblock In \emph{Advances in Neural Information Processing Systems}, volume~32. Curran Associates, Inc.

\bibitem[{Indyk and Motwani(1998)}]{gauss_preserves_norm}
Piotr Indyk and Rajeev Motwani. 1998.
\newblock \href {https://doi.org/10.1145/276698.276876} {Approximate nearest neighbors: towards removing the curse of dimensionality}.
\newblock In \emph{Proceedings of the Thirtieth Annual ACM Symposium on Theory of Computing}, STOC '98, page 604–613, New York, NY, USA. Association for Computing Machinery.

\bibitem[{Li et~al.(2014)Li, Andersen, Park, Smola, Ahmed, Josifovski, Long, Shekita, and Su}]{data_parallel_2}
Mu~Li, David~G. Andersen, Jun~Woo Park, Alexander~J. Smola, Amr Ahmed, Vanja Josifovski, James Long, Eugene~J. Shekita, and Bor-Yiing Su. 2014.
\newblock Scaling distributed machine learning with the parameter server.
\newblock In \emph{Proceedings of the 11th USENIX Conference on Operating Systems Design and Implementation}, OSDI'14, page 583–598, USA. USENIX Association.

\bibitem[{Lialin et~al.(2023)Lialin, Shivagunde, Muckatira, and Rumshisky}]{relora}
Vladislav Lialin, Namrata Shivagunde, Sherin Muckatira, and Anna Rumshisky. 2023.
\newblock \href {https://arxiv.org/abs/2307.05695} {Relora: High-rank training through low-rank updates}.
\newblock \emph{Preprint}, arXiv:2307.05695.

\bibitem[{Liu et~al.(2022)Liu, Zheng, Wang, Cen, Chen, Han, Chen, Liu, Tang, Gonzalez, Mahoney, and Cheung}]{GACT}
Xiaoxuan Liu, Lianmin Zheng, Dequan Wang, Yukuo Cen, Weize Chen, Xu~Han, Jianfei Chen, Zhiyuan Liu, Jie Tang, Joey Gonzalez, Michael Mahoney, and Alvin Cheung. 2022.
\newblock \href {https://arxiv.org/abs/2206.11357} {Gact: Activation compressed training for generic network architectures}.
\newblock \emph{Preprint}, arXiv:2206.11357.

\bibitem[{Liu et~al.(2019)Liu, Ott, Goyal, Du, Joshi, Chen, Levy, Lewis, Zettlemoyer, and Stoyanov}]{roberta}
Yinhan Liu, Myle Ott, Naman Goyal, Jingfei Du, Mandar Joshi, Danqi Chen, Omer Levy, Mike Lewis, Luke Zettlemoyer, and Veselin Stoyanov. 2019.
\newblock \href {https://arxiv.org/abs/1907.11692} {Roberta: A robustly optimized bert pretraining approach}.
\newblock \emph{Preprint}, arXiv:1907.11692.

\bibitem[{Meier and Nakatsukasa(2024)}]{meier2024fast}
Maike Meier and Yuji Nakatsukasa. 2024.
\newblock \href {https://arxiv.org/abs/2105.07388} {Fast randomized numerical rank estimation for numerically low-rank matrices}.
\newblock \emph{Preprint}, arXiv:2105.07388.

\bibitem[{Micikevicius et~al.(2017)Micikevicius, Narang, Alben, Diamos, Elsen, Garcia, Ginsburg, Houston, Kuchaiev, Venkatesh, and Wu}]{mixed_precision}
Paulius Micikevicius, Sharan Narang, Jonah Alben, Gregory Diamos, Erich Elsen, David Garcia, Boris Ginsburg, Michael Houston, Oleksii Kuchaiev, Ganesh Venkatesh, and Hao Wu. 2017.
\newblock \href {https://arxiv.org/abs/1710.03740} {Mixed precision training}.
\newblock \emph{Preprint}, arXiv:1710.03740.

\bibitem[{Miles et~al.(2024)Miles, Reddy, Elezi, and Deng}]{velora}
Roy Miles, Pradyumna Reddy, Ismail Elezi, and Jiankang Deng. 2024.
\newblock \href {https://arxiv.org/abs/2405.17991} {Velora: Memory efficient training using rank-1 sub-token projections}.
\newblock \emph{Preprint}, arXiv:2405.17991.

\bibitem[{Muhamed et~al.(2024)Muhamed, Li, Woodruff, Diab, and Smith}]{grass}
Aashiq Muhamed, Oscar Li, David Woodruff, Mona Diab, and Virginia Smith. 2024.
\newblock \href {https://arxiv.org/abs/2406.17660} {Grass: Compute efficient low-memory llm training with structured sparse gradients}.
\newblock \emph{Preprint}, arXiv:2406.17660.

\bibitem[{Pan et~al.(2022)Pan, Chen, He, Liu, Cai, and Zhuang}]{ACT}
Zizheng Pan, Peng Chen, Haoyu He, Jing Liu, Jianfei Cai, and Bohan Zhuang. 2022.
\newblock \href {https://arxiv.org/abs/2111.11124} {Mesa: A memory-saving training framework for transformers}.
\newblock \emph{Preprint}, arXiv:2111.11124.

\bibitem[{Raffel et~al.(2023{\natexlab{a}})Raffel, Shazeer, Roberts, Lee, Narang, Matena, Zhou, Li, and Liu}]{t5}
Colin Raffel, Noam Shazeer, Adam Roberts, Katherine Lee, Sharan Narang, Michael Matena, Yanqi Zhou, Wei Li, and Peter~J. Liu. 2023{\natexlab{a}}.
\newblock \href {https://arxiv.org/abs/1910.10683} {Exploring the limits of transfer learning with a unified text-to-text transformer}.
\newblock \emph{Preprint}, arXiv:1910.10683.

\bibitem[{Raffel et~al.(2023{\natexlab{b}})Raffel, Shazeer, Roberts, Lee, Narang, Matena, Zhou, Li, and Liu}]{c4}
Colin Raffel, Noam Shazeer, Adam Roberts, Katherine Lee, Sharan Narang, Michael Matena, Yanqi Zhou, Wei Li, and Peter~J. Liu. 2023{\natexlab{b}}.
\newblock \href {https://arxiv.org/abs/1910.10683} {Exploring the limits of transfer learning with a unified text-to-text transformer}.
\newblock \emph{Preprint}, arXiv:1910.10683.

\bibitem[{Shazeer(2020)}]{swiglu}
Noam Shazeer. 2020.
\newblock \href {https://arxiv.org/abs/2002.05202} {Glu variants improve transformer}.
\newblock \emph{Preprint}, arXiv:2002.05202.

\bibitem[{Shoeybi et~al.(2020)Shoeybi, Patwary, Puri, LeGresley, Casper, and Catanzaro}]{tensor_parallelism}
Mohammad Shoeybi, Mostofa Patwary, Raul Puri, Patrick LeGresley, Jared Casper, and Bryan Catanzaro. 2020.
\newblock \href {https://arxiv.org/abs/1909.08053} {Megatron-lm: Training multi-billion parameter language models using model parallelism}.
\newblock \emph{Preprint}, arXiv:1909.08053.

\bibitem[{Touvron et~al.(2023)Touvron, Lavril, Izacard, Martinet, Lachaux, Lacroix, Rozière, Goyal, Hambro, Azhar, Rodriguez, Joulin, Grave, and Lample}]{llama}
Hugo Touvron, Thibaut Lavril, Gautier Izacard, Xavier Martinet, Marie-Anne Lachaux, Timothée Lacroix, Baptiste Rozière, Naman Goyal, Eric Hambro, Faisal Azhar, Aurelien Rodriguez, Armand Joulin, Edouard Grave, and Guillaume Lample. 2023.
\newblock \href {https://arxiv.org/abs/2302.13971} {Llama: Open and efficient foundation language models}.
\newblock \emph{Preprint}, arXiv:2302.13971.

\bibitem[{Wang et~al.(2019)Wang, Singh, Michael, Hill, Levy, and Bowman}]{glue}
Alex Wang, Amanpreet Singh, Julian Michael, Felix Hill, Omer Levy, and Samuel~R. Bowman. 2019.
\newblock \href {https://arxiv.org/abs/1804.07461} {Glue: A multi-task benchmark and analysis platform for natural language understanding}.
\newblock \emph{Preprint}, arXiv:1804.07461.

\bibitem[{Yang et~al.(2024)Yang, Shi, Wang, Zhen, Shi, and Xu}]{approx_bp}
Yuchen Yang, Yingdong Shi, Cheems Wang, Xiantong Zhen, Yuxuan Shi, and Jun Xu. 2024.
\newblock \href {https://arxiv.org/abs/2406.16282} {Reducing fine-tuning memory overhead by approximate and memory-sharing backpropagation}.
\newblock \emph{Preprint}, arXiv:2406.16282.

\bibitem[{Zhao et~al.(2024)Zhao, Zhang, Chen, Wang, Anandkumar, and Tian}]{galore}
Jiawei Zhao, Zhenyu Zhang, Beidi Chen, Zhangyang Wang, Anima Anandkumar, and Yuandong Tian. 2024.
\newblock \href {https://arxiv.org/abs/2403.03507} {Galore: Memory-efficient llm training by gradient low-rank projection}.
\newblock \emph{Preprint}, arXiv:2403.03507.

\end{thebibliography}

\appendix
\newpage
\section{CompAct with SVD}
\label{app:svd}

In this work, random projections are used to compress activations, thereby avoiding the costly SVD step. Performing an SVD on the activation tensor \(x \in \mathbb{R}^{b \times l \times n}\), where \(b\) is the batch size and \(l\) is the sequence length, can introduce considerable overhead since \(b \times l\) is typically large. By contrast, GaLore applies SVD to the gradient \(G \in \mathbb{R}^{n \times m}\), which does not depend on \(b\) or \(l\), making it generally less expensive than compressing activations via SVD.

Nonetheless, one could adopt GaLore’s SVD-based projection within the CompAct framework by running an iteration with uncompressed activations to compute the gradient, from which the SVD projection is derived and then stored for subsequent updates. However, each time the projection matrix is updated, storing the entire activation tensor would significantly increase the GPU memory peak, since the activations typically dominate memory usage. A potential mitigation strategy might involve updating the projection matrix on smaller batches to reduce peak memory requirements. Further exploration of this approach is left for future work.

\section{Pretraining}
\label{sec:appendix_pretrain}

This appendix provides further details about our pre-training experiments. 

\subsection{Hyperparameters}
We adopt the training setup outlined in \cite{galore}, and apply compact to LLaMA-based models of various sizes. Table \ref{tab:llama-sizes} outlines the amount of data and steps used to train the models.

\begin{table}[htpb!]
  \begin{adjustbox}{center,max width=0.95\linewidth}
  \begin{tabular}{c|cc}
    \toprule
     \textbf{Model Size}& \textbf{Steps} &  \textbf{Training Tokens} \\
    \midrule
    60M  & 10K   & 1.3 B \\
    130M & 20K   & 2.6 B \\
    350M & 60K   & 7.8 B \\
    1B   & 100K  & 13.1 B \\
    \bottomrule
  \end{tabular}
  \end{adjustbox}
  \caption{\textbf{Training Step for Llama models. }}
  \label{tab:llama-sizes}
\end{table}

For all the trained models, we use a maximum sequence length of 256, with a batch size of 512. 
We tune the optimal learning rate for each model size from the range  \(1e-3 \leq lr \leq  1e-2\), and choose the best one based on the validation perplexity. We also adopt a learning rate warmup for the first 10\% of the training steps and use a cosine learning rate scheduler that decreases to 0.1 of the initial learning rate.\\
We use a constant scaling factor of \(\alpha = 0.25\).  Additionally, we tuned the projection update, using $T=50$ period T for after searching over [1,10,50,100,200,500,$\infty$] for the LLaMA-130M model, which was then used to train 60M - 350M. For the 1B model, we use $T=200$.

\subsection{CompAct and FlashAttention}
 We applied CompAct to all linear modules within each Transformer block of our LLaMA-based models, except for the output-projection layer in the attention mechanism. FlashAttention \cite{flash_attention}, a fast and efficient implementation of attention, is widely adopted in many architectures, including the models used in this work for both pretraining and finetuning.\\

However, FlashAttention stores its output in memory as part of the activation memory. This same tensor is then fed into the output-projection layer of the attention block, meaning that these two layers share the same activation memory. Consequently, compressing the activation tensor of the attention’s output-projection does not result in overall memory savings, because the FlashAttention output tensor remains in memory. Further, compressing the shared activation between these two layers would introduce errors in the gradient with respect to the FlashAttention input, potentially causing error accumulation across layers. Addressing this issue lies outside the current scope of CompAct.\\
As a result, we left the output-projection layer uncompressed in our experiments. In future work, a custom CUDA kernel for FlashAttention may allow compression of this layer without introducing errors.\\

However, not compressing this layer introduced instability in the pretraining experiments when training with larger learning rates, due to the inconsistent learning rate across different linear layers induced by the scale $\alpha$.
To mitigate this, one can either adjust the learning rate of the output-projection layer by a scale \(\alpha_{out}\), or simply use a smaller learning rate.\\

In our experiments, when training the smaller models  60M-350M we used a large learning rate of \(0.01\) and scaled the learning of the output projection with \(\alpha_{out} = 0.5\alpha\), while for the larger model we simply used a smaller learning rate of \(0.003\).

\subsection{Type Conversion in Normalization Layers}
We note that our implementation of LLaMA's RMSNorm layers did not apply type conversion during pretraining, as we observed that it did not affect model perplexity, but required extra activations. The baseline was measured without type conversion as well making the comparison fair. Hence, all layers were computed in the type of bfloat16 floating point format.

For our finetuning experiments, we presented Roberta-Base, which applies Layer Norm normalization layers rather than RMSNorm, whose default implementations do include type conversions, from bfloat16 to float32 floating point format, although this is very negligible as the effect of these on peak memory is tiny in finetuning.

\section{Fine-tuning}
\label{sec:appendix_finetune}
To be comparable to the results reported in GaLore \cite{galore} as shown in Table \ref{tab:finetune} we report the same metrics as they did, namely, F1 score on QQP and MRPC, Matthew's Correlation for CoLA, Pearson's Correlation for STS-B, and classification accuracy for all others. The numbers reported for GaLore and Baseline are taken from \cite{galore}. We report the \textit{average} performance over three seeds due to the noisy behavior of the training process. All models were trained for 30 epochs with batch size 16, except for CoLA where we used batch size 32 as in GaLore, and a maximum sequence length of 512, a scale $\alpha=2$ was used with $r=8$ and $\alpha=4$ for $r=4$, all with $T=500$. Again, as in GaLore, we searched for a best learning rate per task, searching in \(\{1e-5, 2e-5, 3e-5\}\).
\section{Choice of Projection Matrix - Loss Curve}
\label{sec:appendix_c}
As can be seen below, all different projection types did converge, strengthening the comparison. We can see small spikes in loss when applying VeLoRA's projection every $T=50$ timesteps.

\begin{figure}[!htbp]
    \centering
    \begin{subfigure}{\linewidth}
        \centering
\includegraphics[width=0.9\linewidth]{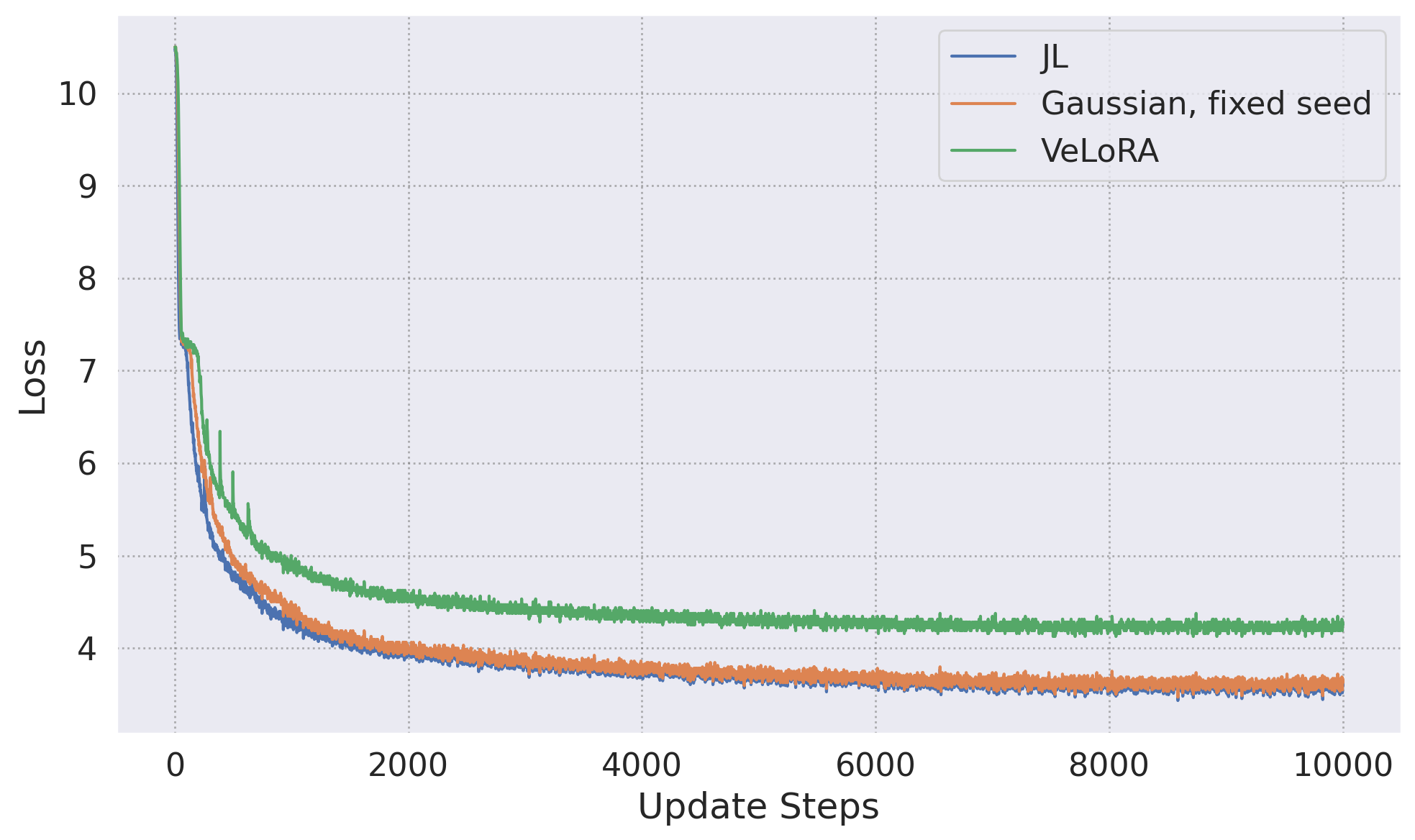}
    \end{subfigure}
    \\[\baselineskip]
    \caption{
     \textbf{Training Loss with $r=n/4$ for different choices of projection matrices.}
     Both Gaussian seed choices and the JL projection achieve comparable results. VeLoRA achieves poor results and is more sensitive to the spike at the beginning of training.
    }
    \label{fig:ab-loss} 
\end{figure}
\newpage
\section{Memory Estimation for Llama Model}
\label{sec:appendix_memory}
In this section, we describe how we calculated the memory estimates used throughout the paper to illustrate CompAct’s memory savings.\\

First, note that given the number of parameters in a model, the memory needed to store gradients and optimizer states can be estimated directly, as it scales with the number of parameters. To estimate activation memory, we examined the activations required by a single Transformer block in a LLaMA-based model as mapped in Table \ref{tab:activations}. We then used this to calculate the total activation memory necessary for training LLaMA models of various sizes.\\

Lastly, we used PyTorch’s memory profiler to confirm that our estimated values are consistent with the actual memory consumption observed during training.

\begin{table*}[htpb!]
\caption{Activations saved from each operation in a single transformer block from a LlaMA model. In this block, the input is of shape \((b, l, n)\), the attention weights \(W_q, W_k, W_v, W_o\) of shape \((n, n)\) and the MLP weights \(W_{down}^\top, W_{up}, W_{gate}\) are of shape \((n, m), n<m\). Note that the activations of RMSNorm don't include the precision conversion as we didn't use it in our training. Smaller activations in CompAct are marked in \textcolor{red}{\textbf{Red}}}
  \begin{adjustbox}{center,max width=0.99\linewidth}
  \begin{tabular}{l|l|l}
    \toprule
     \textbf{Operation}                 &  \textbf{Tensors Saved}       &   \textbf{Tensors Saved With}                 \\
                                        &  \textbf{ for Backward}       &   \textbf{CompAct \((0<=r<=1)\)}              \\
    \midrule
    \(x_2 = RMSNorm(x_1)\)              & \(x_1 \in \R^{b\times l \times n}\)                                           &   \(x_1 \in \R^{b\times l \times n}\) \\
                                        & \((\sigma(x_1)^2 + \varepsilon)^{-\frac{1}{2}} \in \R^{b\times l}\)           &   \((\sigma(x_1)^2 + \varepsilon)^{-\frac{1}{2}} \in \R^{b\times l}\) \\
    \midrule
    \(q = x_2W_q^\top\)                 &                                                                               & \\                                                                 
    \(k = x_2W_k^\top\)                 & \(x_2 \in \R^{b\times l \times n}\)                                           & \textcolor{red}{\((x_2P_q), (x_2P_v), (x_2P_k) \in \R^{b\times l \times nr}\)} \\
    \(v = x_2W_v^\top\)                 &                                                                               & \\
    \midrule
    reshape \(q,k,v\)                   &       None                                                                    & None \\
    to \((b\, h, l, n/h)\)              &                                                                               & \\
    \midrule
    \(x_3 = flash\_attn(q,k,v)\)        &   \(q,k,v \in \R^{b\times h\times l\times n/h}\)                              &   \(q,k,v \in \R^{b\times h\times l\times n/h}\) \\
                                        &   two buffers \(\in \R^{b\times l\times h}\)                                  &   two buffers \(\in \R^{b\times l\times h}\) \\
                                        &   \(x_3 \in \R^{b\times h\times l\times n/h}\)                                &   \(x_3 \in \R^{b\times h\times l\times n/h}\) \\
    \midrule 
    reshape \(x_3\) to \((b, l, n)\)    & None                                                                          & None \\
    \midrule 
    \(x_4 = x_3W_o^\top\)               &   \(x_3 \in \R^{b\times h\times l\times n/h}\)                                &   \(x_3 \in \R^{b\times h\times l\times n/h}\) \\
                                        &   Shared with flash-attn                                                      &   Shared with flash-attn \\
    \midrule
    residual: \(x_5 = x_4 + x_1\)       & None                                                                          & None \\
    \midrule
    \(x_6 = RMSNorm(x_5)\)              & \(x_5 \in \R^{b\times l \times n}\)                                           & \(x_5 \in \R^{b\times l \times n}\) \\
                                        & \((\sigma(x_5)^2 + \varepsilon)^{-\frac{1}{2}} \in \R^{b\times l}\)           & \((\sigma(x_5)^2 + \varepsilon)^{-\frac{1}{2}} \in \R^{b\times l}\) \\
    \midrule
    \(x_{gate} = x_6W_{gate}^\top\)     & \(x_{6} \in \R^{b\times l \times n}\)                                         & \textcolor{red}{\((x_6P_{gate}), (x_{6}P_{up}) \in \R^{b\times l \times nr}\)} \\
    \(x_{up} = x_6W_{up}^\top\)         &                                                                               & \\                                         
    \midrule
    \(x_{act} = SiLU(x_{gate}\)         &  \(x_{gate} \in \R^{b\times l \times m}\)                                     & \(x_{gate} \in \R^{b\times l \times m}\) \\
    \midrule
    \(x_7 = x_{act} * x_{up}\)          &  \(x_{up} \in \R^{b\times l \times m}\)                                       & \(x_{up} \in \R^{b\times l \times m}\) \\
                                        &  \(x_{act} \in \R^{b\times l \times m}\)                                      &  \(x_{act} \in \R^{b\times l \times m}\) \\
    \midrule
    \(x_{down} = x_7W_{down}^\top\)     & \(x_{7} \in \R^{b\times l \times m}\)                                         & \textcolor{red}{\((x_7P_{down}) \in \R^{b\times l \times mr}\)} \\
    \midrule
    residual: \(x_8 = x_down + x_5\)    & None                                                                          & None \\
    \bottomrule
  \end{tabular}
  \end{adjustbox}
  
  \label{tab:activations}
\end{table*}

\end{document}